\documentclass{article} % For LaTeX2e
\usepackage{iclr2023_conference,times}

\usepackage{amsmath,amsfonts,bm}

\def\eqref#1{equation~\ref{#1}}
\def\1{\bm{1}}

\DeclareMathAlphabet{\mathsfit}{\encodingdefault}{\sfdefault}{m}{sl}
\SetMathAlphabet{\mathsfit}{bold}{\encodingdefault}{\sfdefault}{bx}{n}

\usepackage[utf8]{inputenc}
\usepackage{hyperref}
\usepackage{url}
\usepackage{multirow}
\usepackage{multicol}
\hypersetup{
    colorlinks=true,
    linkcolor=cyan
    }
\title{Is multimodal vision supervision beneficial to language?}

\author{Avinash Madasu  \\
Department of Computer Science\\
UNC Chapel Hill, USA \\
\texttt{avinashmadasu17@gmail.com} \\
\And
Vasudev Lal \\
Cognitive Computing Research \\
Intel Labs, 
USA\\
\texttt{vasudev.lal@intel.com}
}

\iclrfinalcopy % Uncomment for camera-ready version, but NOT for submission.
\begin{document}

\maketitle

\begin{abstract}
Vision (image and video) - Language (VL) pre-training is the recent popular paradigm that achieved state-of-the-art results on multi-modal tasks like image-retrieval, video-retrieval, visual question answering etc. These models are trained in an unsupervised way and greatly benefit from the complementary modality supervision. In this paper, we explore if the language representations trained using vision supervision perform better than vanilla language representations on Natural Language Understanding and commonsense reasoning benchmarks. We experiment with a diverse set of image-text models such as ALBEF, BLIP, METER and video-text models like ALPRO, Frozen-in-Time (FiT), VIOLET.  We compare the performance of language representations of stand-alone text encoders of these models to the language representations of text encoders learnt through vision supervision. Our experiments suggest that vanilla language representations show superior performance on most of the tasks. These results shed light on the current drawbacks of the vision-language models. The code is available at \url{https://github.com/avinashsai/MML}
\end{abstract}

\section{Introduction}
Vision-language (VL) pre-training ~\cite{radford2021learning,li2021align,li2022blip,bain2021frozen,fu2021violet} has shown tremendous success in the areas of image-text retrieval ~\cite{li2021align, li2022blip}, visual question answering ~\cite{wang2021simvlm,dou2022empirical}, video retrieval ~\cite{bain2021frozen, fu2021violet, madasu2022learning, madasu2023improving}. These models benefit from the mutual supervision of vision and language leading to the superior results on multi-modal tasks. So, the natural question arises: \textit{``Are vision supervised language representations beneficial compared to vanilla language representations on Natural Language Understanding (NLU) tasks?"} 
To understand this, we conduct a study comparing the language representations trained using only the text to the language representations trained using vision supervision. More specifically, we compare the performance of the text encoders used in vision-language models to the vanilla pre-trained text encoders. 

Few works ~\cite{iki2021effect,singh2022flava} evaluated the performance of vision-language and vanilla language models on GLUE. However, there exists a data discrepancy as these models are pre-trained on different domains of data making the comparisons unfair. To overcome this, we pre-train all the vanilla language models with the text captions used in multi-modal pre-training while keeping the identical training setting. Therefore, the only difference in training between vision-language and vanilla language models is the use of vision data.

For our experiments we use a diverse set of image-text models: ALBEF ~\cite{li2021align}, BLIP ~\cite{li2022blip} and METER ~\cite{dou2022empirical} and video-text models: ALPRO ~\cite{li2022align}, Frozen-in-time (FiT) ~\cite{bain2021frozen} and VIOLET ~\cite{fu2021violet}. We evaluate these models on NLU benchmarks GLUE ~\cite{wang2018glue}, Superglue ~\cite{wang2019superglue} and Common sense reasoning datasets such as SocialIQA ~\cite{sap2019social}, CosmosQA ~\cite{huang2019cosmos}, WinoGrande ~\cite{sakaguchi2021winogrande}, CODAH ~\cite{chen2019codah} and HellaSwag ~\cite{zellers2019hellaswag}. 

Our experiments show that (i) vision supervised language representations under perform compared to vanilla language representations on most of the Natural Language Understanding tasks like Natural Language Inference (NLI), sentence similarity, reading comprehension, linguistic probe and textual entailment. (ii) A similar trend is observed for commonsense reasoning benchmarks.
\section{Related Work}
Over the recent years there has been a tremendous progress in training vision and language together using large-scale multi-modal data. ~\cite{li2019visualbert,chenuniter,li2020oscar}. These models combine both the modalities into a single input and are trained using objectives similar to masked language modelling. Another line of work ~\cite{radford2021learning, li2021align, li2022blip,bain2021frozen} explore dual stream architectures in which there is a separate encoder for each of the modalities and the final representations are minimized using contrastive loss.

Natural Language Understanding involves several tasks such as text classification ~\cite{wang2012baselines, madasu2019sequential}, sentence similarity ~\cite{mueller2016siamese}, Natural Language Inference ~\cite{williams2018broad} etc. However to evaluate the capability of models towards a broad range of NLU tasks, benchmarks such as GLUE ~\cite{wang2018glue}, Superglue ~\cite{wang2019superglue} are introduced. Since then, these benchmarks are being used to comprehensively evaluate the performance of language models.
\begin{table}
\caption{Comparison among different image-text and video-text models in-terms of pre-training data, architecture of the text encoders and size of the text encoder. CC denotes Conceptual captions ~\cite{bain2021frozen}, SBU denotes SBU captions ~\cite{ordonez2011im2text} and VG represents visual genome ~\cite{krishna2017visual}.}
\label{models-pretrain-data}
\begin{center}
\begin{tabular}{|l|l|l|l|l|}
\hline
\bf Type  &\bf Model
&\bf Pre-training Data 
&\bf Text Encoder & \bf Num. layers
\\ \hline \\
\multirow{3}{*}{Image-text} & ALBEF  & CC12M + COCO + SBU + VG (14M) & BERT  & 6 \\
& BLIP  & CC12M + COCO + SBU + VG (14M) & BERT  & 12 \\

& METER & CC3M + SBU + VG (4M) & RoBERTa & 6 \\
\hline
\multirow{3}{*}{Video-text} & ALPRO & CC3M + WebVid-2M (5M) & BERT & 6 \\
& FiT & CC3M + WebVid-2M (5M) &  DistilBERT & 6 \\
& VIOLET & YT180M + CC3M + WebVid-2M (11M) & BERT & 12 \\
\hline
\end{tabular}
\end{center}
\end{table}

\section{Experiments}
\subsection{Models}
We experiment with a diverse set of image-text and video-text models. These models differ in the type of pre-training data used, in the architecture of the text encoder and in the sizes the text encoder. The comparison among the models is shown in the table ~\ref{models-pretrain-data}.
\subsubsection{ALBEF}
ALBEF ~\cite{li2021align} is an image-text model pretrained on conceptual captions 12M (CC12M) ~\cite{sharma2018conceptual}, COCO ~\cite{lin2014microsoft}, SBU captions ~\cite{ordonez2011im2text} and visual genome ~\cite{krishna2017visual}. It's text encoder has a pre-trained BERT ~\cite{kenton2019bert} architecture with six transformer encoder layers.
\subsubsection{BLIP}
BLIP ~\cite{li2022align} is proposed as an extension to ALBEF model pretrained using the same data albeit with a large text encoder. It's text encoder has the same configuration as pre-trained BERT.
\subsubsection{METER}
METER ~\cite{dou2022empirical} is an image-text model pretrained on conceptual captions 3M (CC3M),  SBU captions and visual genome. Pre-trained RoBERTa ~\cite{liu2019roberta} with six transformer encoder layers is used as the text encoder. 
\subsubsection{ALPRO}
ALPRO ~\cite{li2021align} is a video-text model whose text encoder has a pre-trained BERT architecture with six transformer encoders. It is pre-trained on a combined data of conceptual captions 3M (CC3M) and WebVid-2M ~\cite{bain2021frozen}.
\subsubsection{Frozen-in-time (FiT)}
Frozen-in-time ~\cite{bain2021frozen} is a dual stream transformer model pre-trained on both image data  conceptual captions 3M (CC3M) and video data WebVid-2M. DistillBERT ~\cite{sanh2019distilbert} is used as the text encoder.
\subsubsection{VIOLET}
VIOLET ~\cite{fu2021violet} is a multi-modal transformer model pre-trained end-to-end on YouTube 180M (YT180M) ~\cite{zellers2021merlot}, conceptual captions 3M (CC3M) and WebVid-2M. The text encoder follows the BERT architecture.
\subsection{Datasets}
For our analysis, we use GLUE, Superglue and commonsense reasoning datasets such as SocialQA, CosmosQA, WinoGrande, CODAH and HellaSwag. For all these datasets, we evaluate the models on the dev data.
\subsection{Implementation}
For fair comparison between the vision supervised text models and vanilla text models, we pre-train the vanilla text models with the text captions from the datasets used for large scale training of image-text and video-text models. Now, the only difference between these models is the use of vision data. We pre-train vanilla text models in the exact setup as vision-language models. We then fine-tune both the vision supervised text models and vanilla text models on downstream tasks. For GLUE, the maximum sentence length used is 200 and the models are trained for 5 epochs. In case of superglue, 250 is the maximum sentence length and the model are trained for 25 epochs. For commonsense reasoning, the models are trained for 10 epochs and 300 is the maximum sentence length. Unless otherwise stated, the results reported are the average of 5 runs.

\begin{table}[!h]
\caption{Results on GLUE benchmark. MNLI-mis refers to the task MNLI mismatched and WNLI denotes the Winograd Schema Challenge. We see that language representations learnt through vision supervision under performs compared to vanilla language representations on all the tasks except WNLI.}
\label{glue}
\begin{center}
\begin{tabular}{|l|l|l|l|l|l|l|l|l|l|}
\hline
  \bf Model &\bf Type &\bf
MNLI &\bf MNLI-mis	&\bf QQP	&\bf SST2	&\bf MRPC &\bf	CoLA &\bf	RTE &\bf	WNLI \\
\hline
\multirow{2}{*}{ALBEF} & Text & 82.77 & 82.68 & 90.54 & 91.44 & 72.81 & 81.50 & 58.12 & 46.01 \\
& Image-text & 61.38 & 61.68 & 79.02 & 80.39 & 66.49 & 69.13 & 50.30 & 56.34 \\
\hline
\multirow{2}{*}{BLIP} & Text & 83.04 & 82.70 & 90.54 & 91.44 & 72.81 & 81.50 & 58.12 & 46.01 \\
 & Image-text & 61.38 & 61.68 & 79.02 & 80.39 & 66.49 & 69.13 & 50.30 & 56.34 \\
\hline
\multirow{2}{*}{METER} & Text & 86.59 & 86.15 & 90.99 & 93.27 & 76.06 & 82.58 & 64.02 & 56.34 \\
 & Image-text & 31.82 & 31.82 & 77.91 & 81.12 & 66.49 & 69.13 & 47.29 & 56.34 \\
\hline
\hline
\multirow{2}{*}{ALPRO} & Text & 82.96 & 82.81 & 90.64 & 92.05 & 70.96 & 79.93 & 60.41 & 45.07 \\
 & Video-text & 62.53 & 63.26 & 79.35 & 80.96 & 66.49 & 69.13 & 54.39 & 56.34 \\
 \hline
 \multirow{2}{*}{FiT} & Text & 79.10 & 80.23 & 89.51 & 52.03 & 72.58 & 69.13 & 57.28 & 48.83 \\
 & Video-text & 59.54 & 59.45 & 79.01 & 52.18 & 66.78 & 69.13 & 48.01 & 56.34 \\
 \hline
 \multirow{2}{*}{VIOLET} & Text & 83.19 & 83.59 & 90.68 & 92.74 & 71.92 & 81.66 & 59.93 & 52.58 \\
 & Video-text & 61.38 & 61.68 & 79.02 & 80.39 & 66.49 & 69.13 & 50.30 & 56.34 \\
\hline
\end{tabular}
\end{center}
\end{table}

\begin{table}[!h]
\caption{Results on Superglue benchmark. WiC represents Word-in-Context, CB represents CommitmentBank, COPA denotes Choice of Plausible Alternatives and WSC means The Winograd Schema Challenge.}
\label{superglue}
\begin{center}
\begin{tabular}{|l|l|l|l|l|l|l|}
\hline
\bf Model &	\bf Type &	\bf BoolQ & \bf	WiC	& \bf CB	& \bf COPA & \bf WSC \\
\hline
\multirow{2}{*}{ALBEF} & Text & 70.41 & 63.13 & 76.79 & 48.00 & 63.46 \\
 & Image-text & 63.30 & 55.02 & 63.93 & 51.60 & 63.46 \\
\hline
\multirow{2}{*}{BLIP} & Text & 70.41 & 63.13 & 76.43 & 48.00 & 63.46 \\
 & Image-text & 63.30 & 55.02 & 63.93 & 51.60 & 63.46 \\
\hline
\multirow{2}{*}{METER} & Text & 72.40 & 66.11 & 75.00 & 46.80 & 63.46 \\
 & Image-text & 66.87 & 53.98 & 69.64 & 50.80 & 63.46 \\
\hline
\hline
\multirow{2}{*}{ALPRO} & Text & 71.16 & 67.18 & 76.79 & 42.20 & 63.46 \\
 & Video-text & 65.17 & 53.17 & 62.50 & 50.60 & 62.50 \\
\hline
\multirow{2}{*}{FiT} & Text & 68.91 & 62.38 & 69.29 & 44.80 & 63.46 \\
 & Video-text & 64.69 & 53.20 & 70.71 & 53.80 & 63.46 \\
\hline
\multirow{2}{*}{VIOLET} & Text & 63.85 & 57.37 & 66.07 & 56.00 & 63.46 \\
 & Video-text & 63.44 & 54.11 & 63.93 & 52.60 & 63.46 \\
\hline

\end{tabular}
\end{center}
\end{table}

\begin{table}[!h]
\caption{Results on Commonsense reasoning tasks.}
\label{commonsense}
\begin{center}
\begin{tabular}{|l|l|l|l|l|l|l|}
\hline
\bf Model & \bf	Type &\bf SocialQA &\bf	CosmosQA & \bf WinoGrande & \bf	CODAH &\bf HellaSwag \\
\hline
\multirow{2}{*}{ALBEF} & Text & 40.50 & 26.45 & 53.12 & 25.72 & 25.04 \\
 & Image-text & 33.47 & 25.24 & 49.57 & 25.72 & 24.48 \\
\hline
\multirow{2}{*}{BLIP} & Text & 52.27 & 25.72 & 56.88 & 26.02 & 25.24 \\
 & Image-text & 33.47 & 25.24 & 49.57 & 25.72 & 24.48 \\
\hline
\multirow{2}{*}{METER} & Text & 58.39 & 31.32 & 59.59 & 24.40 & 25.04 \\
 & Image-text & 33.47 & 25.00 & 49.57 & 25.72 & 24.48 \\
\hline
\hline
\multirow{2}{*}{ALPRO} & Text & 49.90 & 27.45 & 56.56 & 24.10 & 24.89 \\
 & Video-text & 33.96 & 25.70 & 50.28 & 25.72 & 24.48 \\
\hline
\multirow{2}{*}{FiT} & Text & 45.46 & 30.87 & 56.75 & 25.12 & 26.54 \\
 & Video-text & 33.35 & 25.77 & 50.33 & 24.52 & 24.59 \\
\hline
\multirow{2}{*}{VIOLET} & Text & 43.36 & 33.17 & 57.09 & 24.28 & 25.27 \\
 & Video-text & 33.47 & 25.24 & 49.57 & 25.72 & 24.48 \\
\hline
\end{tabular}
\end{center}
\end{table}

\section{Results}
Table ~\ref{glue} shows the results on GLUE benchmark. From the tables, it is evident that vanilla language representations show superior performance compared to vision supervised language representations on most of the tasks across all the models. The drop in performance is significant for NLI tasks like MNLI and MNLI-mismatched (MNLI-mis). A similar trend is observed for sentence similarity (QQP), sentiment classification (SST2), reading comprehension (MRPC), linguistic probe (CoLA) and textual entailment (RTE). However, we see a huge improvement in performance for the Winograd NLI (WNLI) task.

Table ~\ref{superglue} illustrates the results on superglue benchmark. From the table, we observe that vision supervised language representations under perform compared to vanilla language representations. For the tasks question answering (BoolQ), word in context (WiC), discourse (CB) we see a huge drop in performance. However, we see a significant improvement in performance for the casual reasoning (COPA) task. It is worth-noting that the performance is same for both the vanilla and vision supervised language representations on winograd schema challenge (WSC).

Table ~\ref{commonsense} demonstrates the results on commonsense reasoning datasets. As shown in the table, the performance of vanilla language representations surpass vision supervised language representations. There is a notable difference in performance on SocialQA, CosmosQA, WinoGrande and HellaSwag commonsense tasks. However for the CODA dataset, we observe vision supervised language representations outperform vanilla language representations for METER, ALPRO and VIOLET models.

\section{CONCLUSION AND FUTURE DIRECTIONS}
In this paper we comprehensively evaluated if the vision supervised language representations are beneficial to the language. We experimented with three image-text models ALBEF, BLIP, METER and three video-text models ALPRO, FiT, VIOLET on NLU benchmarks GLUE, superglue and commonsense reasoning tasks. Our experiments showed that vanilla language representations significantly outperform vision supervised language representations on most of the tasks. We believe these findings can shed light on the future directions to improve the vision-language pre-training that is beneficial to understanding the language. 
\bibliography{iclr2023_conference}
\bibliographystyle{iclr2023_conference}

\end{document}